\documentclass[graybox]{svmult}

\usepackage[pages=all, color=black, position={current page.south}, placement=bottom, scale=1, opacity=1, vshift=5mm]{background}
\SetBgContents{

}      

\usepackage[margin=1in]{geometry} 

\usepackage{amsmath}
\usepackage{amssymb}
\usepackage{algorithm2e}
\usepackage[utf8]{inputenc}
\title*{Offline Handwritten Mathematical Recognition using  Adversarial Learning and Transformers  }	

\author{Ujjwal Thakur$^1$ , Anuj Sharma$^1$ \newline Department of Computer Science and Applications,
Panjab University, Chandigarh, India }

\begin{document}

\maketitle

\abstract { Offline Handwritten Mathematical Expression Recognition (HMER) is a major area in the field of mathematical expression recognition. Offline HMER is often viewed as much harder problem as compared to online HMER due to lack of temporal information and variability in writing style. In this paper, we purpose a encoder-decoder model that uses paired adversarial learning. Semantic-invariant features are extracted from handwritten mathematical expression images and their printed mathematical expression counterpart in the encoder. Learning of semantic-invariant features combined with DenseNet encoder and transformer decoder, helped us to improve the expression rate from previous studies. Evaluated on the CROHME dataset, we have been able to improve latest CROHME 2019 test set results by 4\% approx.
}


  \section{Introduction}\label{intro}
Digital information resources are easy to access, search and distribute as compared to traditional printed documents. This makes the availability of existing paper books and articles in the digital form necessary, as the major population now has access to inexpensive digital devices. Automated techniques to 
convert traditional printed documents into computer encoded form have received researchers attention for a long time now \cite{Chan2000} \cite{Memon} \cite{Drobac} \cite{wei} \cite{Zhang}.   Optical Character Recognition (OCR) has made huge accuracy and reliability improvements in recent years \cite{Neudecker}. The OCR commonly recognizes 1-D linear structure while Mathematical Expressions Recognition (MER) works on symbols with 2D spatial relationship \cite{YanZuo}, and with variable scale. This makes MER harder as compared to OCR and, as a result, makes scientific books and documents that include mathematical expressions have a low recognition rate as compared to plain text documents \cite{HeFukeng}. This combined with mathematical expressions playing important role in engineering disciplines has attracted special attention to MER field in recent years \cite{ZHANG2017196}. This particular pattern recognition problem usually has two types as online and offline expressions \cite{ZheDm}. The online expressions are sequences of points in space and retention time information of the strokes used to write the expression \cite{Zhele}. While offline formulas are represented as images \cite{ALVARO2016135}. Our focus in this paper will be on offline expressions. \par
One aspect of the problem related to MER is Handwritten Mathematical Expression Recognition (HMER). The HMER has larger variability as compared to MER as individuals have a unique way to write characters and symbols \cite{ariva}. This makes the HMER problem complex as compared to printed mathematical expression recognition. The HMER have the application of writing handwritten mathematical expression in computer encoded form without using additional tools such as \LaTeX. As Keyboards do not have enough keys to input mathematical symbols, writing mathematical formulas into digital documents becomes a time-consuming and tiresome task. Tools such as MS Word and latex are available, but they come with an additional learning curve. If HMER techniques can predict handwritten expressions with a low error rate, these can save precious time for educators, researchers and mathematicians. It is a challenging task since math symbols have a 2-dimensional layout with ABOVE, LEFT UP, BELOW, CONTAIN, SUPERSCRIPT, HORIZONTAL properties \cite{ZhaXi}. Symbols that have similar visual properties such as 'y' and '$\gamma$', 'p' and '$\rho$' also make it hard for the recognizer to differentiate \cite{YanZuo}. The HMER has two major steps as symbol recognition and structural analysis \cite{Chan2000} \cite{Zanibbi2012}. These steps can be completed in a sequential or in a parallel manner. When the sequential approach is used, the expressions are first divided into individual math symbols and recognized \cite{Zanibbi2012}. In the structural analysis step, best results from the previous step are taken to analyze the structure of the mathematical expression \cite{Alavro2014}. The Sequential approaches have a problem that errors introduced in previous steps can affect the results of the next step \cite{Alvaro2011}. While methods with parallel approaches go through steps simultaneously and have been used extensively in the past. But parallel methods become complex as they are working with global information, this makes our training and learning models resource-hungry.
 \par
 Abstractly, our motive is to take images of handwritten expressions and output a 1-dimensional linear latex script for that expression. There are mainly two kinds of techniques used by researchers for the recognition of the mathematical expression. First set of techniques have grammar and automata at their core while second set of techniques have artificial intelligence and deep learning at their core \cite{Sakshi}. Grammar-based approaches need to manually define grammar rules which can be a tedious task and the number of rules increases as we try to make our method more accurate. They also fail when we introduce some expression that does not fit the listed grammar rules. In recent years, researchers have shifted their focus to methods based on artificial intelligence and deep learning \cite{Sakshi}. The methods based on deep learning have given better results than grammar-based methods  \cite{Sakshi}, and are better prepared for an unknown type of expression input.
 \par

 Deep learning techniques have recently adopted encoder-decoder models for HMER problem \cite{Zhang2018MultiScaleAW} \cite{ZHANG2017196}. The encoder-decoder models have been used by researchers in other fields as, image segmentation \cite{Badrina},dialogues generation \cite{Serban}, object detection \cite{Yuzhu}, image restoration \cite{MaoXi}. In encoder-decoder models, we pass input through the encoder and convert that input into a required higher form. Then the output of the encoder is given to the decoder for further processing. The major advantage of using this type of model is that, we can have variable size inputs that can be converted by the encoder to a fixed-length vector and passed to the decoder. In this paper, we have also used encoder-decoder architecture with a DenseNet model encoder and transformer model as the decoder. We use the encoder to extract semantic-invariant from handwritten expression images in presence of printed images of the same expression and a Bi-directional transformer as the decoder.

The organization of the rest of this paper is as follows. The section 2 presents a literature review and discusses 
conventional approaches, deep learning-based approaches in HMER.  The section 3 discusses the proposed model and section 4 presents experimental results and comparison. In section 5, we discuss of different aspects of the model during experiments. Finally, section 6 concludes this paper.

\section{Related work}\label{Related work}
In the past, many approaches have been proposed to solve the HMER problem \cite{ALVARO2016135}\cite{ZHANG2017196}\cite{YanZuo}\cite{Zanibbi2012}\cite{Alavro2014}\cite{Zhang2018MultiScaleAW}. These approaches can be categorized mainly into three categories as, grammar-based, deep learning-based and miscellaneous. We consider those approaches into miscellaneous, that make use of rule based approaches, statistical approach \cite{Celik} and few other \cite{Simis}. The previous researchers mostly processed symbol recognition, structural analysis separately and used rule-based methods. Grammar-based approaches became prevalent later on, these were able to do steps in parallel. Various predefined grammars are used by researchers to solve HMER such as stochastic context-free grammars, relational grammars, and definite clause grammars \cite{ZhaoWe}. In Anh Duc Le et al. research ME are represented in the form of stochastic context-free grammar (SCFG), the Cocke–Younger–Kasami (CYK) algorithm is used to parse two-dimensional (2D) structures of online handwritten ME. Then CYK select the best interpretation from the results of symbol segmentation, symbol recognition and structural analysis. Grammar-based approaches were the top choice of researchers before deep learning emerged as a better alternative. One of the advantage of  deep learning techniques is that, researcher need not to manually write grammar rules for different types of structures.

Deep learning models with encoder-decoder have achieved favorable results in the recent past. The various deep learning techniques have been used by the researchers such as support vector machine \cite{LeAn}, random forest, recurrent neural network \cite{Endong}, convolution neural network \cite{YanZuo}.
The deep learning methods are different from methods based on predefined grammar and heuristic rules in the aspect that deep learning methods at their core are driven by learning data. This makes methods with deep learning easy to train on a wide range of symbols and expressions. 
In some aspects, mathematical expression image recognition can be categorized as a sequence-to-sequence problem. Jianshu Zhang et al. \cite{ZHANG2017196} used a CNN as encoder and RNN with attention as a decoder in their model called Watch, Attend, Parse (WAP). The encoder-decoder models have one major advantage as they can have variable-length input and output sequences \cite{Sutskever2014}. In the training phase, to predict the next expression, the decoder is provided with expressions and their ground truth label as input. In the prediction phase, the model utilizes a beam search to choose the most likely candidates for the next predicted expression \cite{ZHANG2017196}. When dealing with an image-based sequence problem, Baoguang Shi et al. proposed an encoder based on deep CNN and RNN. The RNN is used because of its capability to capture contextual information within a sequence. These models generally extract abstract features that are an abstract representation of the input image. Later on, these features are decoded to generate the context sequence. The encoder-decoder is jointly trained to make training coordinated and model well-optimized \cite{ZHANG2017196}.
 Although in some cases encoder is pre-trained on a specific set of data to make it accurate in feature extraction \cite{XuKe}. 
Jin-Wen Wu et al. \cite{Wu2020HandwrittenME} applied paired adversarial learning for HMER. An encoder-decoder model is used for the guidance of the printed image expression. The encoder DenseNet CNN is used as a feature extractor with a densely connected multi-directional RNN block. The decoder uses attention while decoding. Recognizer and discriminator are trained to differentiate between printed and handwritten expression images to make it better at extracting semantic features. Zhao et al. also used encoder-decoder architecture with CNN model as encoder and transformer model as a decoder. The transformer is trained using bi-directional language modeling \cite{ZhaoWe}.

Ashish Vaswan et al. made a big step in the use of the attention mechanism \cite[]{Vaswani}. His research became the core mechanism for the transformer model. Transformers overcame shortcomings of the RNN models, when it came to sequence to sequence problems with longer sequences. Older models failed to retain previous information when sequences become too long. Attention works by taking the weighted sum of previous encoder states. This allows the decoder to assign larger weights to important inputs when mapping to elements in the output. Self-attention is used by the transformer to find out how important are other input elements with respect to the current input element. Recurrent units are not used in tranformers, this makes it suitable for parallel computations while extracting features.

\section{Proposed Model}
In this section, we propose a model based on the encoder-decoder framework for HMER. The encoder is equipped to learn semantic invariant features of the mathematical expressions. The decoder have transformer with attention and positional encoding. This model takes an image of mathematical expression as an input and output equivalent latex code. \par

 \begin{tikzpicture}
  \node[inner sep=0pt] (image) at (0,0) {\includegraphics[scale=0.05]{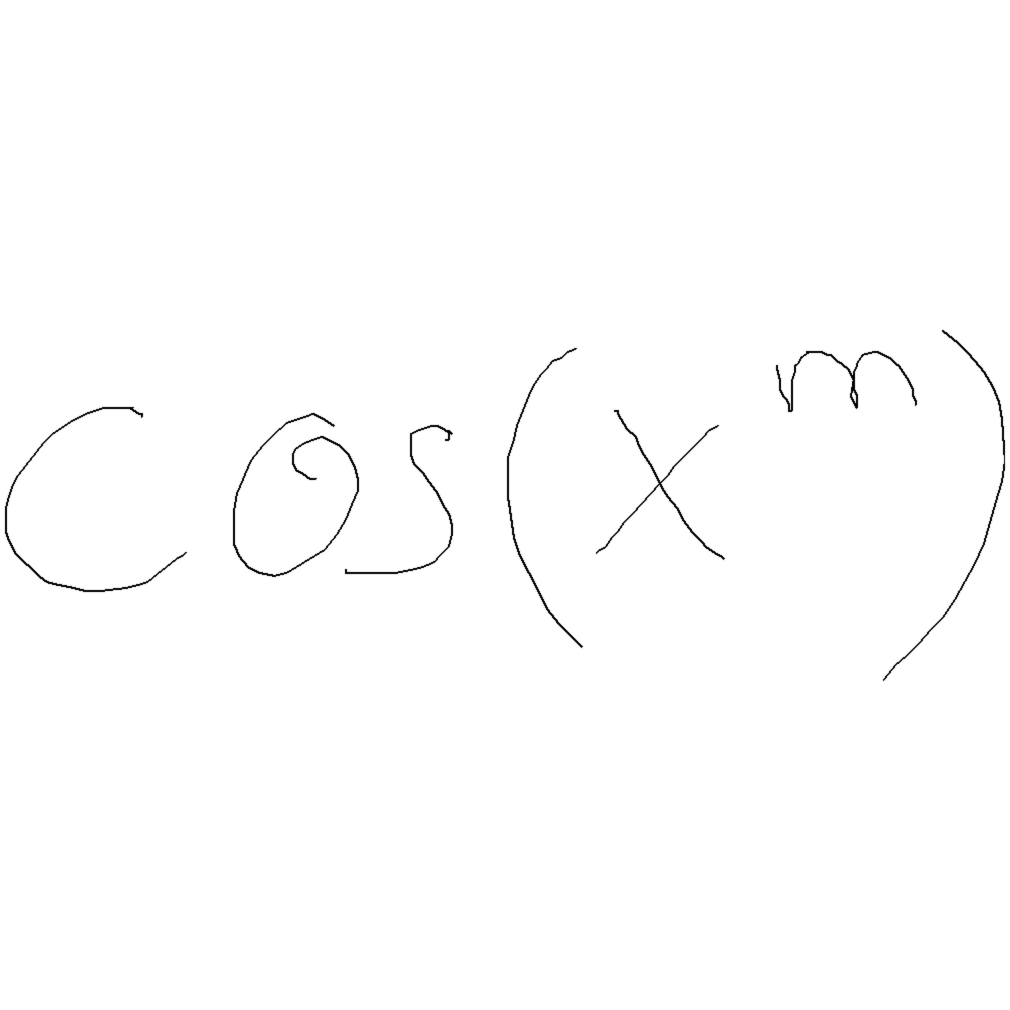}};
  \node[rectangle,draw] (model) at (2,0) {Model};
  \node[rectangle,draw] (tex_script) at (4,0) {cos(x\textasciicircum m) };
  \draw[->,thick](image)--(model);
  \draw[->,thick] (model)--(tex_script); 
\end{tikzpicture}

In this work, we utilize the methodology proposed by Jin-Wen Wu et al. \cite{Wu2020HandwrittenME} to learn semantic-invariant features from the handwritten mathematical symbols. We have mainly four components as recognizer, discriminator, enocder and decoder. Discriminator helps recognizer in the extraction of semantic invariant features. This makes the recognizer have better accuracy in the case of writing style variation. Printed images and handwritten images are given as input to the model in a paired manner. Recognizer is trained to recognize handwritten and printed images while extracting such features that our discriminator is unable to distinguish between features of a handwritten and printed expression image. Recognizer and discriminator are continuously updated during training.

\subsection{Learning Semantic-Invariant Features}
{
  Traditionally, models are trained to learn features that are discriminative to differentiate between symbols and classes. This decreases their accuracy when dealing with the writing style variations of different individuals. Adversarial feature learning pairs printed symbols with their handwritten expression and the recognizer is trained to learn semantic-invariant features in the presence of the discriminator. Attention is used while pairing the handwritten and printed symbols. To define this process clearly, if we have  $d\left(a, b, \theta_{R}\right)=\left(d_{1}, \ldots, d_{L}\right) \in \mathbb{R}^{L \times C}$, $ \left(d_{1}, \ldots, d_{L}\right)$ is sequence of features from most recent decoder block. $a$ is the input image which can be handwritten or printed, $b=\left(b_{0}, \ldots, b_{L-1}\right)$ is previously recognized expression and  $\theta_{R}$ are recognizer parameters. The recognizer is guided by discriminator $D$ to learn semantic-invariant by deciding whether the vector $d_l$ containing image features, is from handwritten or printed ME. The $d_l$ is from a printed ME template, this probability can be calculated by \[D\left(d_{l}\left(a_{p}, b_{0: l-1}, \theta_{R}\right), \theta_{D}\right)\]

where, $b_{0: t-1} = b_{0}, \ldots, b_{t-1}$, $a_p$ is printed ME template and $\theta_{D}$ are the discriminator features.

Then  function 
\[
  \begin{aligned}
\begin{split}
\mathcal{P}_{D}=& E_{(A, B)} E_{l}\left(\log D\left(a_{l}\left(a_{p}, b_{0: l-1}, \theta_{R}\right), \theta_{D}\right)\right. \\
&+\log \left(1-D\left(a_{l}\left(a_{h}, b_{0: l-1}, \theta_{R}\right), \theta_{D}\right)\right)
\end{split}
\end{aligned}
\]

Here, $l \in\{1, \ldots, L\}$, and $(A, B)=\left\{\left(\left(a_{h}, a_{p}\right), b\right)\right\}$ are set of paired images, handwritten and printed ME images are used for training. The discriminator $D$ is optimized to maximize the probability of correctly classifying whether features source is handwritten or printed. We can state $D$ is optimized to maximize $\mathcal{P}_D$. The recognizer $R$ is optimized to learn semantic-invariant features to fool discriminator. This optimization problem can be regarded as minimizing the loss that $d_{l}$ is classified from a printed image.

This optimization problem can be written as 

$$
\mathcal{P}_{D_{\text {adv }}}=-E_{(A, B)} E_{l}\left(\log D\left(d_{l}\left(a_{h}, b_{0: l-1}, \theta_{R}, \theta_{D}\right)\right)\right.
$$

The main goal of our recognizer is to correctly predict symbols in the input handwritten mathematical expression image. The $d_l$ which output feature needs to be classified as $b_l$ with high probability at each decoding step. The function to classify features that are learned using handwritten ME images can be written as 

$$
\mathcal{P}_{C_{h}}=-E_{\left(A_{h}, B\right)} \sum_{l=1}^{L} \log p\left(b_{l} \mid a_{h} ; b_{0: l-1} ; \theta_{R}\right)
$$
where, $\left(A_{h}, B\right)=\left\{\left(a_{h}, b\right)\right\}$ is the set of training images with handwritten images.  $p\left(b_{l} \mid a_{h} ; b_{0: l-1} ; \theta_{R}\right)$ is the probability of $b_{l}$ getting correctly classified after applying last step, softmax function.

The loss function for the features learned from the printed templates of ME can be written as, 
$$
\mathcal{P}_{C_{p}}=-E_{\left(A_{p}, B\right)} \sum_{l=1}^{L} \log p\left(b_{l} \mid a_{p} ; b_{0: l-1} ; \theta_{R}\right)
$$
where $\left(A_{p}, B\right)=\left\{\left(a_{p}, b\right)\right\}$ are the pairs of training set made up of printed templates.

The final loss function that needs to be minimized can be written as, 
$$
\mathcal{P}_{R}=\mathcal{P}_{C_{h}}+\mathcal{P}_{C_{p}}+\delta \mathcal{P}_{D_{\text {adv }}},
$$

Where, $\delta$ hyperparameter keeps the balance between discriminative and semantic-invariant feature learning. The $\delta$ and semantic-invariant features learning are directly proportional which intern makes discriminative feature learning inversely proportional to $\delta$. This is done by extracting fewer features of the respective type.
}

\begin{figure*}[h]
\hfill  \includegraphics[width=10cm,scale=0.1]{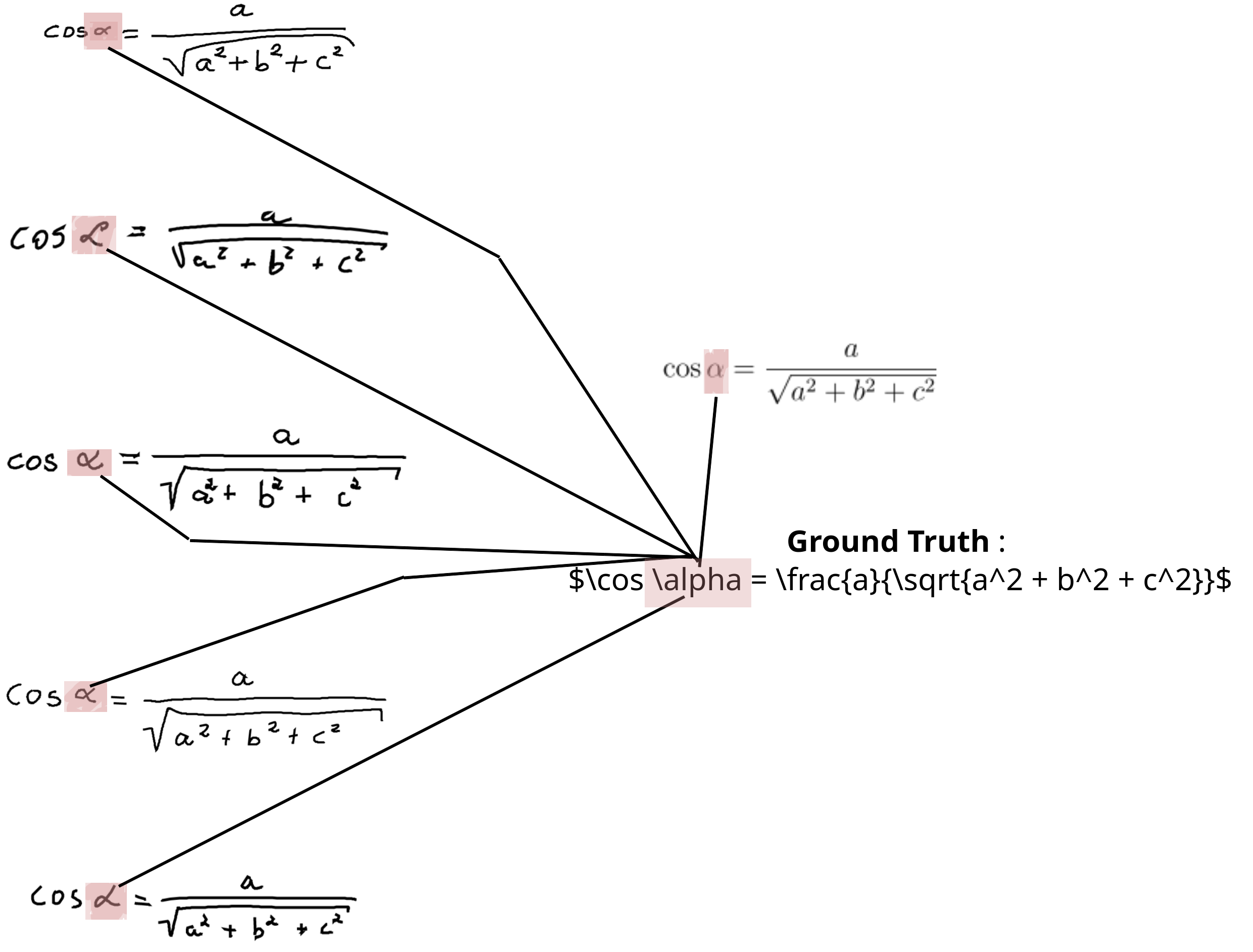} \hspace*{\fill}
  \caption{Visualizing how images are paired. }
  \label{fig:mapping}  
\end{figure*}

In the figure \ref{fig:mapping} ,  MEs are written by different individuals, on the right top, we have the printed template and on the right bottom, we have the ground truth in latex code. The highlighted symbols in the image represent the attention area with high symbol probability. The lines are drawn to show that the same symbol can be written in very different ways. Here, all of the styles share some invariant features while representing identical semantic meaning.

In the encoder, DenseNet is used as feature extractor in the encoder part. The DenseNet \cite{DensNet} is a suitable for our encoder part as it uses concatenation to combine feature-maps of all the previous layers. If we have $0^{th}$ to $l^{th}$ layers with output features $z_0,z_1.......z_l$ then $(l+1)^{th}$ layer can written as   \[z_(l+1) = Q_{l+1}\left(\left[x_{0}, x_{1}, \ldots, x_{l}\right]\right)\]. Here, $\left[z_{0}, z_{1}, \ldots, z_{l}\right]$ is concatenation operation on all the output feature maps of previous layers. The $Q_{l+1}$ is function to denote operation of three layers, a batch normalization, followed by ReLu layer and convolutional layer of size $3 \times 3$. As collective information is passed from all the previous layers, DenseNet is compact and efficient in computation as compare to other alternatives like ResNet, AlexNet, VGGNet and enables better propagation of gradient.Researchers have found out that DenseNet performed better with fewer parameters as comapred to ResNet which another popular deep learning model \cite[]{Hekain}.  \par

\begin{figure*}[h]
  \hfill \includegraphics[width=8cm,scale=0.1]{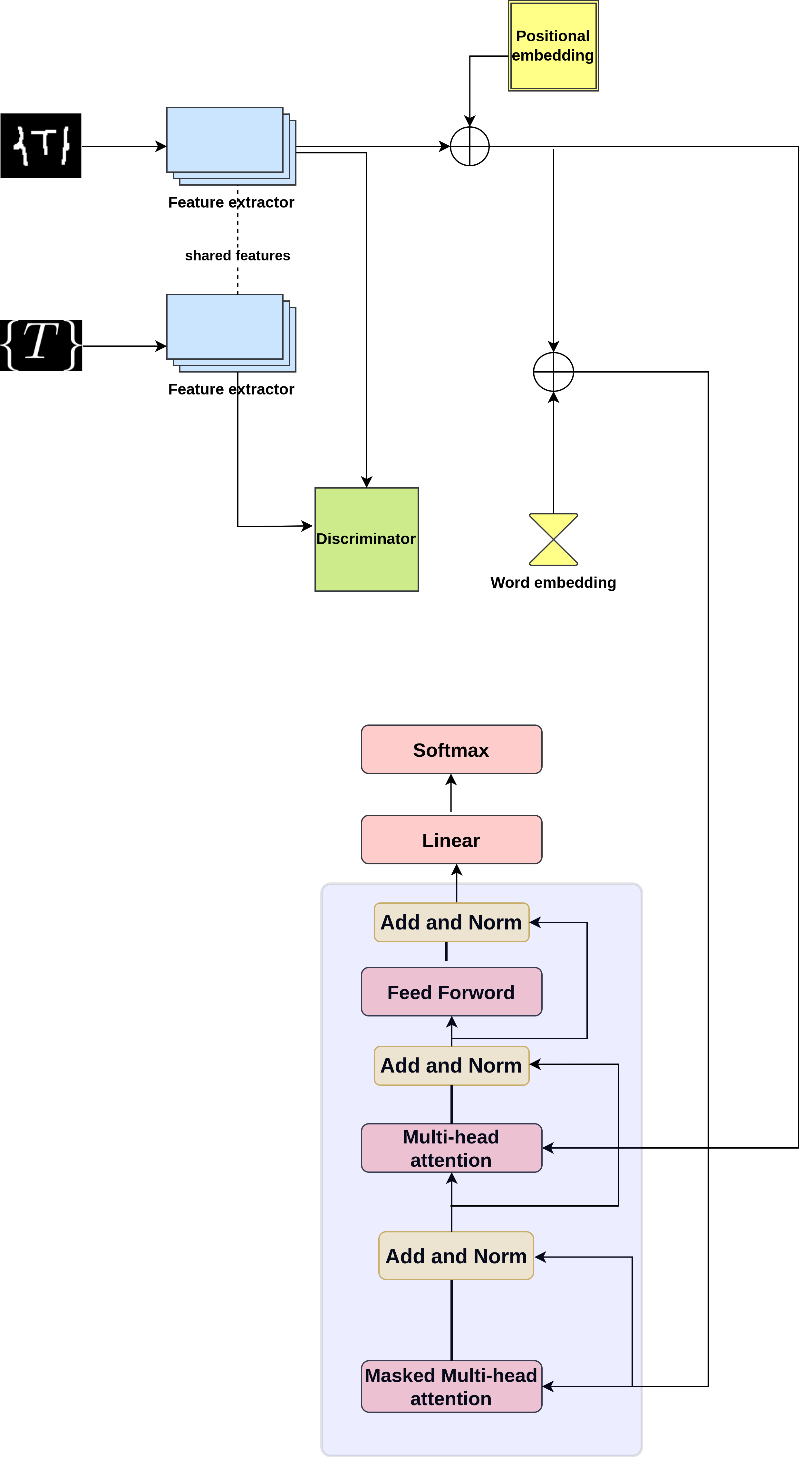} \hspace*{\fill}
  \caption{Outline of the model}
  \label{fig:modeldia}
\end{figure*}

The proposed outline of model has been presented in figure \ref{fig:modeldia}. The model start by giving the paired image of the printed template and handwritten mathematical expression image as the input to the feature extractor. These features are shared to fool the discriminator and help the recognizer to improve its accuracy. These learnings are then used to make feature extractor better at extracting semantic-invariant features. Positional embedding and word-level embeddings are done on the extracted features before sending them to the transformer decoder. Masked multi-head attention and multi-head attention are used to focus on the parts of the image that need more focus. The decoder is composed of three normalization layers, one feed foreword layer, one linear layer and one softmax layer at the end. The decoder is then trained on the training data set with attention to the semantic-invariant features extracted by the feature extractor in the presence of the recognizer and discriminator.

During decoding as each data point is treated as independent in the transformers, it helps in the parallelization of the computations, but the order of words and their position is also important information. Word positional encoding and image positional encoding are used. This information help model to attend regions that are important in the input image. In previous studies \cite{ZhaoWe}, where RNN is used, RNN takes the order of the words or symbols into consideration but neglected information regarding the position of the features in the image. In positional encoding, a unique value is assigned to describe the position of the object in the sequence. In our case, objects are features of the image. If we have a feature at $x$ position, then $\vec{p}_x  \in \mathbb{R}^{d} $ will be its encoding where $d$ be dimensions of encoding. Two types of encoding are used word positional encoding and image positional encoding. Word positional encoding is defined by sine and cosine functions.

Let's suppose we have a feature at $x^{th}$ position then encoding is done in following way, where $Y : \mathbb{N} \rightarrow \mathbb{R}^{d} $  , \par

\[ Y (x,2i,d,n) = \sin(x/n^{2i/d})   \]
\[ Y (x,2i+1,d,n) = \cos(x/n^{2i/d})   \]
\\

where, $x$ is the given position, $d$ is the dimension of the output matrix, $n$ is user-defined, and $i$ is an index in dimension. We have used $ n$ as 10000 as in the paper ``Attention is all you need'' \cite{Vaswani}. We can write  $\vec{p}_x$  as a vector with pairs of sines and cosines for each data point.

$$
\overrightarrow{Y_{x}}=\left[\begin{array}{c}
\sin \left(\omega_{1} \cdot x\right) \\
\cos \left(\omega_{1} \cdot x\right) \\ \\
\sin \left(\omega_{2} \cdot x\right) \\
\cos \left(\omega_{2} \cdot x\right) \\ \\
\vdots \\ \\
\sin \left(\omega_{d / 2} \cdot x\right) \\
\cos \left(\omega_{d / 2} \cdot x\right)
                             \end{array}\right]_{d \times 1}
$$

  where, $ \omega_{i} = 1 / 10000^{2i/d}, d$ is encoding dimension.
Another encoding that is used is Image positional Encoding. Image positional features are represented with the help of 2-D normalized positional encoding. Similar to word encoding, we compute pairs of sinusoidal positional functions of two dimensions and later concatenate them. When we have positional tuple $(x,y)$ and dimension d as the word encoding. Image positional encoding vector is represented as

  \[P^I_{x,y,d} = [P^W_{\bar{x},d/2}; P^W_{\bar{y},d/2}    ]  \]

  where $\bar{x} = x / H$,  $\bar{y} = y / W$, H is Height of image and W is width of the image. After performing the above steps we have all the encoding information. Now comes the decoding part, where, we will be using the transformer model. There are four basic parts in each decoder layer module. \\

\textbf{Scaler Dot-product Attention}  The $\sqrt{d_{k}}$ is used to scale down the dot products. Here $ {d_{k}}$ is dimensions. If there is a  query $Q$, a key $K$ and value $V$ then attention can be written as,

$$
\operatorname{Attention}(Q, K, V)=\operatorname{softmax}\left(\frac{Q K^{T}}{\sqrt{d_{k}}}\right) V
$$

The attention mechanism works in manner that, whenever, there is a query, it fetches the value from key-value pairs. The decision of, what value is to be fetched is made on the bases of similarity between the query and the key. We can compute our output matrix in parallel here. Reason to divide by $\sqrt{d_{k}}$ is that if we assume $Q$ and $K$ are vectors with $d_k$ dimensions that have random variables with $0$ mean and $1$ variance. Their dot product $Q \cdot K = \sum_{i=1}^{d_k} u_iv_i$ where u and v are values from the $Q$ and $K$ vectors. This dot product has a mean 0 and variance $d_k$.  We prefer these values to have variance 1.

\textbf{Multi-Head Attention}  Computations are repeated by the attention module in parallel multiple times. The module doing these computations is called Attention Head. The query, key and value parameters are split into smaller batches by attention module and each batch passes through a different head independently. All these similar computation results are combined to make a final attention score. As multiple modules are used to get a single score called multi-head attention. Multi-head attention allows multiple relationships and refinements for each word to be encoded into a single score.

If we have matrices with projection parameters $\mathbf{W}_{i}^{Q} \in \mathbb{R}^{d_{\text {model }} \times d_{Q}}, \mathbf{W}_{i}^{K} \in$ $\mathbb{R}^{d_{\text {model }} \times d_{k}}, \mathbf{W}_{i}^{V} \in \mathbb{R}^{d_{\text {model }} \times d_{v}}$, $\mathbf{H}_{i}$ for  the batch of query $\mathbf{Q}$, key $\mathbf{K}$, and value $\mathbf{V}$ .
$$
\mathbf{H}_{i}=\operatorname{Attention}\left(\mathbf{Q} \mathbf{W}_{i}^{Q}, \mathbf{K} \mathbf{W}_{i}^{K}, \mathbf{V} \mathbf{W}_{i}^{V}\right)
$$

All the $h$ number of  heads are concatenated and then projected with a $\mathbf{W}^{O} \in \mathbb{R}^{h d_{v} \times d_{\text {model }}}$  parameter matrix.

$$
\operatorname{MH}(\mathbf{Q}, \mathbf{K}, \mathbf{V})=\left[\mathbf{H}_{1} ; \ldots ; \mathbf{H}_{h}\right] \mathbf{W}^{O}
$$

\begin{figure}
  \includegraphics[width=73mm,scale=0.4]{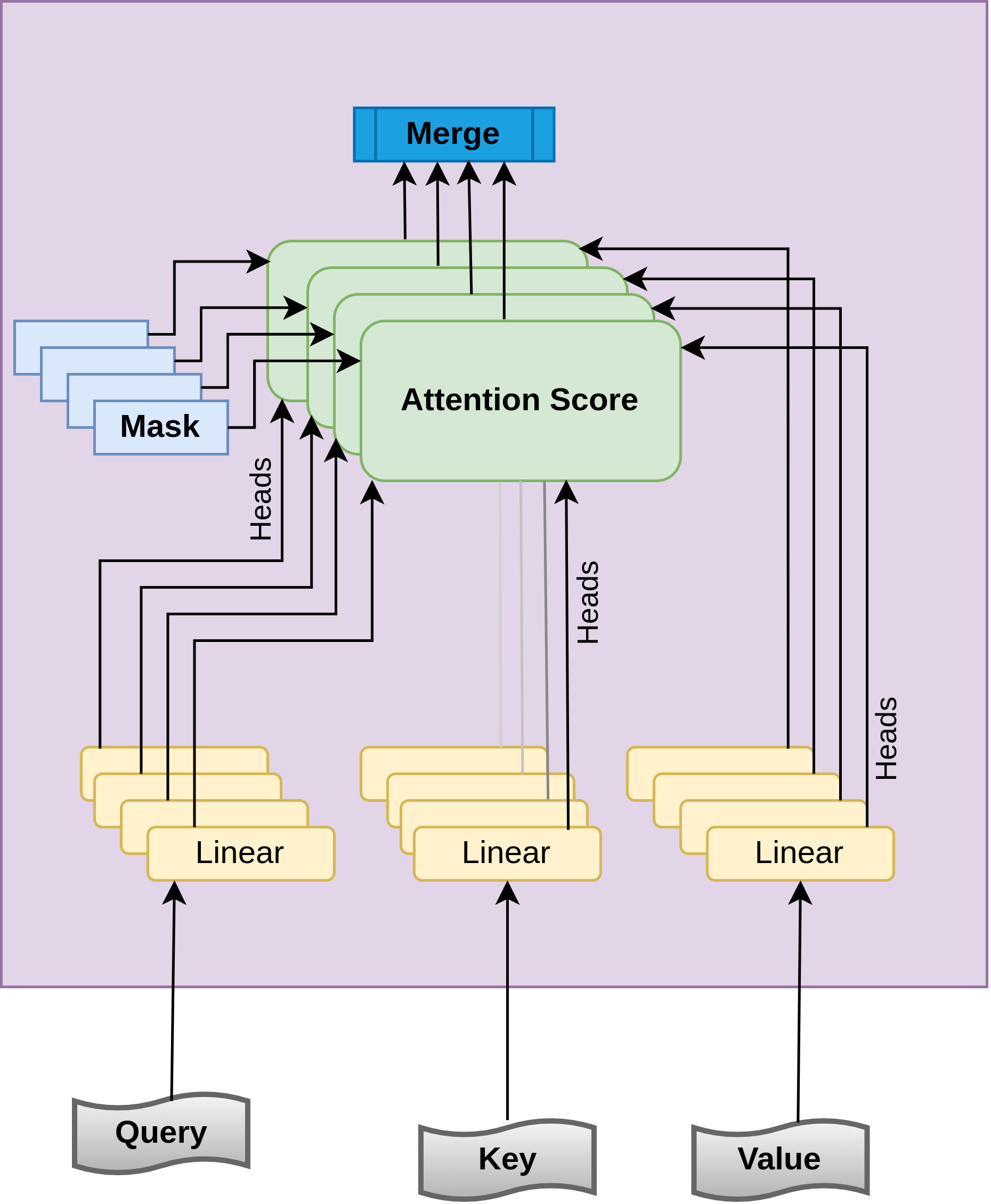}
 \caption{Working of attention}
  \label{fig:attention}
\end{figure}

In the figure  \ref{fig:attention}, the attention layer is taking three input parameters value, key, and query. These three have a similar structure as a single vector is used to represent each symbol. These are first passed through the linear layer as we can see in the figure. After passing through the linear layer a matrix representation is produced for each of value, key and query. The produced matrix is given as input to the attention layer. As we have discussed earlier in multi-head attention data is split and then feed into multiple attention layers this can be seen in the figure visualized with multiple attention layers and multiple inputs given to them by the linear layer. This is merely a logical separation. Physically, the Query, Key, and Value are not separated into independent matrices, one for each Attention head. For the Query, Key, and Value, a single data matrix is employed, with logically independent regions of the matrix for each Attention head. Every Attention head doesn't have an individual Linear layer either. The Attention heads all work on the same Linear layer, but on their own logical segment of the data matrix.

\textbf{Masked Multi-Head Attention} masked multi-head attention mechanism during the training process only one forward computation. As masking makes training parallel. While decoding because of autoregressive property, the input image and previously generated symbols make the base for the prediction of the next symbol. During training, the mask matrix enables the self-attention module to restrict the attention region for each time step.

\textbf{Position-wise Feed-Forward Network} In the decoder part feed-forward network does three operations: Linear transformation, Activation based on ReLU and linear transformation.

$FFN(x) = max(0, xW_1 + b_1 ) W_2 + b_2$

  Feed foreword network enables internal information of each position to be integrated separately.
  
\begin{algorithm}

  \SetAlgoLined

\vspace{0.7cm}
  
\KwIn{Paired images of printed template and handwritten mathematical expression.}
\KwResult{Trained model using adversarial learning and tranformer decoder.}
 Get printed mathematical expression $a_{p}$ for $a_{h}$ by compiling its label $b$ to obtain the training set $\left(\left(a_{h}, a_{p}\right), b\right) \in(A, B)$\; 
 Initialize the recognizer model and the discriminator randomly with parameters $\theta_{R}$ and $\theta_{D}$\;

 \While{$\mathcal{P}_{C_{h}}+\mathcal{P}_{C_{p}}+\delta \mathcal{P}_{D_{a d v}}$ is not converged}{
Sample $n$ pairs of samples $\left\{\left(a_{h}, a_{p}\right)^{(1)}, \ldots,\left(a_{h}, a_{p}\right)^{n}\right\}$ from the training set; Where n is batch size\;

Update the recognizer by
$\theta_{R} \leftarrow \theta_{R}+$ optim $\left(-\frac{\partial\left(\mathcal{P}_{C_{h}}+\mathcal{P}_{C_{p}}+\delta \mathcal{P}_{D_{a d v}}\right)}{\partial \theta_{R}}, \eta_{R}\right)$\;

\For {m steps}{

Update the discriminator by
$\theta_{D} \leftarrow \theta_{D}+\operatorname{optim}\left(\frac{\partial \mathcal{P}_{D}}{\partial \theta_{D}}, \eta_{D}\right) \vspace{1mm}$ \;
 }
 }

Parameterize the recognizer by $\theta_{R}$\;
Train recognizer transformer with attention\;
return The recognizer $R$\;

\caption{Algorithm of the model}
\end{algorithm}

In the Algorithm 1, we have explained the process step by step. First, we take pairs of printed templates and handwritten images from the training data set. We initialize parameters for the recognizer and discriminator, while training we feature extractor and discriminator against each other to perform accurately. To start this task, recognizer and discriminator are initialized with random parameters. In the next step, we start the process of training the recognizer. We update recognizer parameters $\theta_R$ by minimizing the loss of classifying features learned from handwritten images, and printed images and the loss that current features are classified from the printed images. \par 

In the same way, we repeatedly update the discriminator parameters by optimizing the discriminator loss function for a predecided number of times. We repeatedly keep updating both discriminator and recognizer parameters until the above-described loss functions are converged. Then transformer decoder 

\section{Experiments}
The proposed model is trained and tested on the dataset available from the Competition on Recognition of
Online Handwritten Mathematical Expressions (CROHME) \cite{Mouchre2016AdvancingTS}. We have extensively tested each part of the proposed model and compared it with other state-of-the-art approaches by other researchers. 

\subsection{Datasets and Training}
We have used Competition on Recognition of Online Handwritten Mathematical Expressions, which largest publicly available dataset for handwritten mathematical expression. The training set has $8836$ handwritten mathematical expressions. The test set of CROHME has 986 images in the 2014 competition, 1147 in the 2016 competition, and 1199 in the 2019 competition. \par
In the competition, every handwritten mathematical expression is saved in an InkML file. Each InkML file contains hand stroke trajectory information and ground truth in MathML and Latex formats. We take these InkML files and convert that trajectory information to offline bitmap image formate for training and testing purposes. CROHME has provided official evaluation tools and ground truth.

\begin{figure*}[h]
  \includegraphics[width=17cm,scale=0.3]{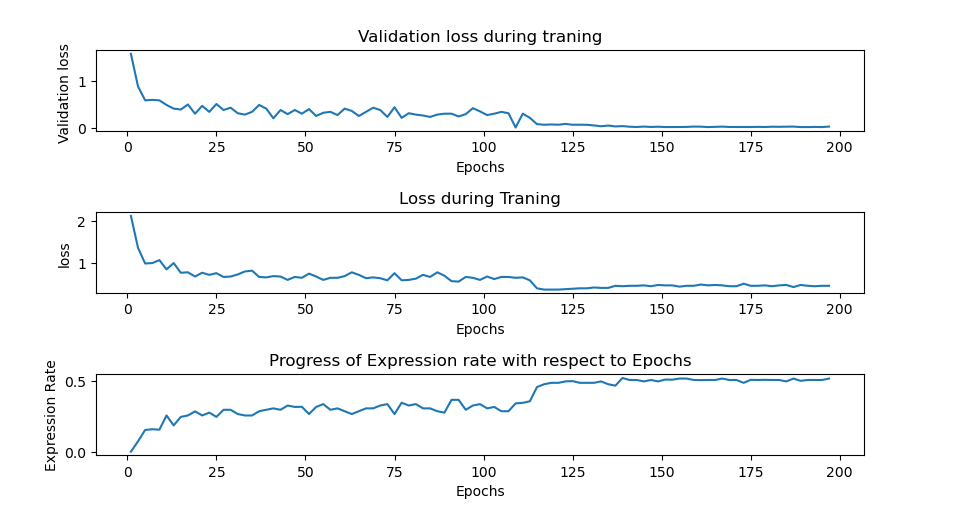}
  \caption{Loss, Validation loss and expression rate during training }
  \label{fig:graphs}
\end{figure*}

During training, we train the model using the training images given by the CROHME dataset and printed versions of the same mathematical expression. From Fig \ref{fig:graphs}, we can see that loss and validation loss both decreases gradually as the number of epochs increases. Expression rate of the trained model increases gradually as the number of epochs increases. 

\subsection{Compare with state-of-the art results}

The results shown in the table compare models on the CROHME 2014,2016 and 2019 datasets. To keep comparison fair models use only the official provided dataset of 8836 training images. We have not used any data argumentation techniques or other methods. For CROHME 2019 official methods, we use “Univ. Linz” \cite{univ}  method as the baseline. For Image-to-L A TEX methods, we use the previous state-of-the-art
“WYGIWYS” \cite{Deng2016WhatYG}, “PAL-v2” \cite{Wu2020HandwrittenME} , “WAP” \cite{ZHANG2017196} , “Weakly supervised WAP”(WS WAP) \cite{10.1007/978-3-031-02375-0_2}, “DenseWAP”  \cite{Zhang2018MultiScaleAW}  and the tree decoder-based “DenseWAP-TD” \cite{pmlrv119zhang20g} method. The “Ours”  methods denote the model trained using the  transformer decoder and adversarial learning.

\begin{figure}

\[\begin{array}{ccc}
 \text { Test Dataset } & \text { Model } & \text { exp rate (\%) } \\
\hline  \\
  \text{CROHME19}& \text { Univ. Linz } \cite{univ} & 41.49 \\
& \text { DenseWAP } \cite{Zhang2018MultiScaleAW} & 41.7 \\
& \text { DenseWAP-TD } \cite{pmlrv119zhang20g} & 51.4 \\
& \text { Tran-Uni } \cite{ZhaoWe} & 44.95 \\
& \text { Tran-Bi } \cite{ZhaoWe} & {52 . 96} \\
& \textbf{ Our Method }  & \textbf{56.38} \\

  \hline \\
 {\text { CROHME16 }} & \text { WAP } \cite{ZHANG2017196} & 37.1 \\
& \text { DenseWAP } \cite{Zhang2018MultiScaleAW} & 40.1\\ 
& \text { PAL-v2 } \cite{Wu2020HandwrittenME} & 49.61 \\
 & \text { DenseWAP-TD } \cite{pmlrv119zhang20g} & 48.5 \\
& \text { WS WAP } \cite{10.1007/978-3-031-02375-0_2} & 51.96 \\
& \text { Tran-Uni } \cite{ZhaoWe} & 44.55 \\
                         & \text { Tran-Bi } \cite{ZhaoWe} & \textbf{52.31} \\
    & \textbf{ Our Method } \cite{ZhaoWe} & \textbf{52.31} \\

  \hline\\
{\text { CROHME14 }}& \text { WYGIWYS }\cite{Deng2016WhatYG} & 36.4 \\
& \text { DenseWAP } \cite{Zhang2018MultiScaleAW} & 43.0 \\
& \text { PAL-v2 } \cite{Wu2020HandwrittenME} & 48.88 \\
& \text { DenseWAP-TD } \cite{pmlrv119zhang20g} & 49.1\\ 
& \text { WS WAP } \cite{10.1007/978-3-031-02375-0_2} & 53.65 \\
& \text { Tran-Uni } \cite{ZhaoWe} & 48.17 \\
 & \text { Tran-Bi } \cite{ZhaoWe} & \textbf{53.96} \\
 & \textbf{ Our Method } & {51.67} \\

  \hline
\end{array}\]

  \caption{Comparison of expression rate with other models }
\end{figure}

\section{Discussion}
\subsection{Hyperparameter $\mathcal{P}$ effect on model}
The hyperparameter determines the trade-off between semantic-invariant characteristics and the recognizer's learned discriminative features. When its value is small, discriminative features account for the majority of the recognizer's loss and dominate back-propagated gradients. The recognizer learns more semantic-invariant aspects of the same symbols in handwritten ME pictures and their printed templates as an increase in the value of the hyperparameter. However, if the value is too large, the model will be too focused on learning semantic-invariant characteristics and to fool the discriminator it will try creating the identical feature sequences for printed and handwritten ME pictures. As a result, various symbol groups will have fewer distinguishing traits, and identification accuracy will suffer. In a worst-case scenario, the recognizer might only focus on sections such as background and other less important regions at each cycle to fool the discriminator. Due to its importance in training the model, choosing a suitable value for the hyperparameter is an important step. 

\subsection{Analysing the discriminator }
 Using a discriminator to help the recognizer learn semantic invariant characteristics improves recognition. A discriminator with a bigger capacity performs better. It is preferable to pay attention to individual symbols in ME pictures with complex 2D structures rather than using global average pooling to obtain the image vector. Even though a discriminator with a larger capacity improves performance, we found that powerful discriminators are more prone to gradient disappearance during training. We set the discriminator and classifier to the same number of layers to guarantee that the training process is stable.

\subsection{ Effect of Attention}
MEs with longer target sequences are more likely to have wider widths and more complex structures in general. The recognizer model is prone to over and under attention as a result of these MEs. It is noticed that when attention is used, the identification accuracy on both test sets improves for most target length intervals. More surprisingly, after using attention on the CROHME 2014 test set, some handwritten MEs with over 40 symbols were successfully recognized. To put it another way, by using attention, the model can better attention to the key areas in decoding.

\subsection{Comparison of results}
We provide various case studies for the "DenseWAP," "Tran-Uni," and "tran-Bi" models. The DenseNet encoder is shared by these three models. These three models differ in that "DenseWAP" employs an RNN-based decoder, "Trans-Uni" adapts a standard transformer decoder, and "Tran-Bi" trains a transformer decoder using a bidirectional training method.
First off, it is clear from comparing the predictions made by "DenseWAP" and "Tran- Uni" that the "DenseWAP" model is unable to accurately anticipate all of the symbols in input pictures with complex structure. Ours model was able to beat the state of the art results in 2019 CROHME test with 56.38\% expression rate.

\section{Conclusion }
This work uses a paired adversarial learning strategy in combination with the transformers to solve the problem of handwritten mathematical expression recognition. By learning both semantic-invariant and discriminative characteristics, the proposed model, dubbed outperforms others in coping with writing-style variance. Furthermore, our model can successfully read 2D spatial structures thanks to the attention mechanism, despite the training ME pictures having only weak labels in LaTeX format.
We show that the proposed outperforms state-of-the-art approaches on the recent public datasets CROHME 2019 and this suggest that paired adversarial learning method and attention are beneficial in improving performance through extensive trials and studies.

	\bibliography{sn-article}

\end{document}